\documentclass[conference]{IEEEtran}
\IEEEoverridecommandlockouts
\usepackage{cite}
\usepackage{amsmath,amssymb,amsfonts}
\usepackage{algorithmic}
\usepackage{booktabs}
\usepackage{graphicx}
\usepackage{textcomp}
\usepackage{xcolor}
\usepackage{multirow}
\usepackage{algorithm}
\usepackage{multirow}
\usepackage{amsmath}
\usepackage{hyperref}
\usepackage{balance}
\usepackage{threeparttable}

\def\BibTeX{{\rm B\kern-.05em{\sc i\kern-.025em b}\kern-.08em
    T\kern-.1667em\lower.7ex\hbox{E}\kern-.125emX}}

\newcommand{\methodname}{MMCircuitEval}
    
\begin{document}

\title{\methodname: A Comprehensive Multimodal Circuit-Focused Benchmark for Evaluating LLMs}

\author{
\IEEEauthorblockN{
Chenchen Zhao$^{*1,7}$\thanks{$^*$Equal contribution},
Zhengyuan Shi$^{*1,7}$,
Xiangyu Wen$^{*1,7}$,
Chengjie Liu$^{2,7}$,
Yi Liu$^{1,7}$,\\
Yunhao Zhou$^{1,7}$,
Yuxiang Zhao$^{3,7}$,
Hefei Feng$^{4,7}$,
Yinan Zhu$^{7}$,
Gwok-Waa Wan$^{7}$,\\
Xin Cheng$^{5,7}$,
Weiyu Chen$^{2,7}$,
Yongqi Fu$^{4,7}$,
Chujie Chen$^{6,7}$,
Chenhao Xue$^{3,7}$,\\
Guangyu Sun$^{3}$,
Ying Wang$^{6}$,
Yibo Lin$^{3}$,
Jun Yang$^{4,7}$,
Ning Xu$^{5,7}$,
Xi Wang$^{4,7}$, and
Qiang Xu$^{+1,7}$\thanks{$^+$Corresponding author: Qiang Xu (qxu@cse.cuhk.edu.hk)}
}
\IEEEauthorblockA{$^1$\textit{Department of Computer Science and Engineering, The Chinese University of Hong Kong}}
\IEEEauthorblockA{$^2$\textit{School of Electronic Science and Engineering, Nanjing University}}
\IEEEauthorblockA{$^3$\textit{School of Integrated Circuits, Peking University}}
\IEEEauthorblockA{$^4$\textit{School of Intergrated Circuits, Southeast University}}
\IEEEauthorblockA{$^5$\textit{School of Computer Science and Engineering, Southeast University}}
\IEEEauthorblockA{$^6$\textit{Department of Computer Science and Technology, University of Chinese Academy of Sciences}}
\IEEEauthorblockA{$^7$\textit{National Center of Technology Innovation for EDA}}
}

\maketitle

\begin{abstract}
The emergence of multimodal large language models (MLLMs) presents promising opportunities for automation and enhancement in Electronic Design Automation (EDA). However, comprehensively evaluating these models in circuit design remains challenging due to the narrow scope of existing benchmarks. To bridge this gap, we introduce \methodname, the first multimodal benchmark specifically designed to assess MLLM performance comprehensively across diverse EDA tasks. \methodname~comprises 3614 meticulously curated question-answer (QA) pairs spanning digital and analog circuits across critical EDA stages—ranging from general knowledge and specifications to front-end and back-end design. Derived from textbooks, technical question banks, datasheets, and real-world documentation, each QA pair undergoes rigorous expert review for accuracy and relevance. Our benchmark uniquely categorizes questions by design stage, circuit type, tested abilities (knowledge, comprehension, reasoning, computation), and difficulty level, enabling detailed analysis of model capabilities and limitations. Extensive evaluations reveal significant performance gaps among existing LLMs, particularly in back-end design and complex computations, highlighting the critical need for targeted training datasets and modeling approaches. \methodname~provides a foundational resource for advancing MLLMs in EDA, facilitating their integration into real-world circuit design workflows. Our benchmark is available at \url{https://github.com/cure-lab/MMCircuitEval}.

\end{abstract}

\vspace{-10pt}
\section{Introduction} \label{sec:intro}
In recent years, large language models (LLMs) have showcased remarkable capabilities across various domains, effectively automating tasks and enhancing productivity in fields such as natural language processing, software engineering, and data analysis. Recognizing this potential, the semiconductor industry has started exploring the role of LLMs in circuit design. Recent initiatives, such as ChipNemo~\cite{liu2023chipnemo} and SemiKong~\cite{semikong2024}, underscore the promise of LLMs in assisting engineers with circuit analysis and design optimization within Electronic Design Automation (EDA) workflows.

To evaluate the efficiency and effectiveness of LLMs in the circuit design domain, researchers have developed various circuit-focused benchmarks aimed at assessing these models from diverse perspectives. Some existing benchmarks~\cite{liu2024openllm,lu2024rtllm,qiu2024autobench,liu2023verilogeval, li2025deepcircuitx} concentrate on assessing the quality of Verilog code snippets generated by LLMs. While these benchmarks provide a range of design specifications paired with corresponding implementations or testbenches, the designs are generally far smaller and less complex than those encountered in practical scenarios. Other benchmarks~\cite{wu2024eda,pu2024customized} attempt to evaluate LLMs based on their abilities to select EDA tools and generate design flow scripts. However, these narrowly focused tasks limit the scope of tool planning and leave unanswered questions about whether LLMs truly understand the intricate circuit features. Consequently, comprehensively evaluating LLMs for circuit design remains a critical challenge. 

To address this gap, we introduce \methodname: a comprehensive multimodal circuit-focused benchmark specifically designed to evaluate LLM performance across various stages and types of circuit designs. \methodname~comprises 3,614 question-answer (QA) pairs collected from diverse and reliable sources, including open-source \textit{textbooks}, technical \textit{question banks}, and \textit{online resources}. To enhance realism and practicality, we also generate additional questions derived from datasheets, register-transfer level (RTL) codes, and netlists of \textit{real-world products}, covering a wide range of circuit-related challenges. Each QA pair undergoes meticulous manual review by domain experts to ensure it meets high standards of accuracy, relevance, and technical depth. 

\methodname~is structured to offer a dual perspective, catering to both circuit designers and LLM developers. On the one hand, \methodname~systematically categorizes QA pairs based on circuit types (digital and analog) and distinct stages of EDA workflows (general knowledge, design specification, front-end design, and back-end design). This organization allows for a detailed evaluation of how effectively LLMs can assist hardware engineers at different stages of circuit design. On the other hand, \methodname~includes detailed metadata for each QA pair, specifying the abilities being tested (domain-specific knowledge, multimodal comprehension, logical reasoning, and numerical computation) and difficulty levels (easy, medium, and hard). This detailed labeling not only provides granular insights into model performance but also serves as a guide for optimizing LLM design to better support complex circuit design from the perspective of LLM developers.


We conduct extensive experiments to evaluate a variety of common LLMs using our proposed benchmark. Experimental results reveal that most widely-used LLMs fail to achieve satisfactory performance in circuit-focused question answering. Notably, the models struggle the most with \textit{back-end design} and \textit{circuit-related computations}, highlighting the urgent need for relevant training corpora and processing techniques tailored to circuit-focused LLM training. We also discuss potential solutions to address these challenges, including illustrative experiments to validate their impacts.

To the best of our knowledge, \methodname~is the first benchmark designed to assess LLM capabilities across various multimodal circuit-related questions and different EDA stages. We hope that \methodname~will serve as a foundational resource, inspiring further innovation in leveraging LLMs to develop advanced solutions to EDA challenges.

\begin{figure*} [!t]
    \centering
    \includegraphics[width=1.0\linewidth]{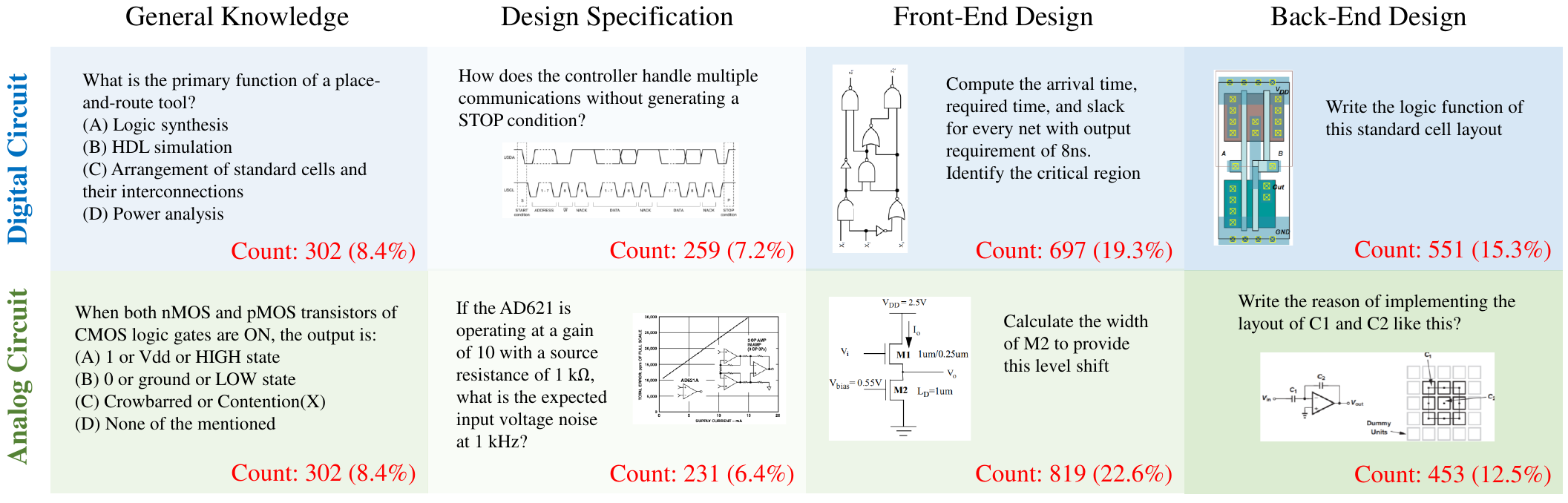}
    \caption{The overview of \methodname~and sampled questions. Better viewed with zoom-in.}
    \label{fig:enter-label}
\end{figure*}

\section{Related Work} \label{sec:relatedwork}

\subsection{LLM for EDA}
The rapid advancements of AI and LLMs have expanded their applications to various professional domains, including EDA~\cite{chen2024dawn, li2022deepgate, shi2023deepgate2, shi2024deepgate3, zheng2025deepgate4}. Specifically, LLMs have demonstrated significant potential for enhancing EDA workflows in areas such as hardware code generation and verification, EDA tool planning, and interactive question answering systems. For instance, specific LLM-based approaches~\cite{thakur2023autochip, liu2024rtlcoder, thakur2024verigen, liu2024ampagent} focus on automatically generating circuit designs, while others are employed to plan and manage EDA tools~\cite{wu2024chateda, ajayi2019openroad} (e.g.,~automating task decomposition, script generation, and task execution within the EDA design flow). Interactive systems such as ChipGPT~\cite{chang2023chipgpt} and ChipNemo~\cite{liu2023chipnemo} further showcase the utility of LLMs for question answering (QA) and knowledge retrieval, enabling engineers to access relevant design information and troubleshoot issues more effectively. While these advancements highlight the versatility of LLMs in EDA, there remains a notable lack of general and comprehensive benchmarks to evaluate LLM performance across a wide range of EDA tasks.


\subsection{Multimodal Benchmarks for LLM Evaluation}
Recent research has developed a variety of multimodal benchmarks~\cite{li2023seed, yue2024mmmu,liu2025mmbench} to quantitatively measure the performance of LLMs across diverse use cases. These benchmarks conduct extensive evaluations of domain knowledge capacity, comprehension, and logical inference of LLMs in a wide range of professional vision-language scenarios such as mathematics, medicine, art, and engineering. Model performance is quantitatively represented by the average correctness of answers to multimodal questions of different types (e.g.,~choice, open-ended, etc), tested capabilities, and downstream scenarios.


In the domain of circuit design and EDA, existing benchmarks for evaluating LLMs remain limited in narrow scopes. For instance, studies~\cite{liu2024openllm, qiu2024autobench, liu2023verilogeval} focus primarily on assessing LLM capabilities in generating hardware description code and testbenches. Meanwhile, Wu et al.~\cite{wu2024eda} introduces an open-source dataset centered around the OpenROAD~\cite{ajayi2019openroad} EDA toolchain, comprising question-answer pairs, code snippets, and their corresponding OpenROAD scripts. Similarly, another benchmark~\cite{pu2024customized} offers high-quality question-document-answer triplets that address various queries within the EDA design flow. However, these benchmarks primarily focus on applying EDA tools based on documentation rather than understanding design methodologies or analyzing specific designs. More recently, Xu et al. propose ChatICD-Bench~\cite{xu2024chipexpert}, a comprehensive benchmark spanning multiple subfields of chip design. While promising, ChatICD-Bench still lacks comprehensive performance analyses and offers limited question diversity.

\section{The \methodname~Benchmark} \label{sec:method}

\subsection{Overview}

The MultiModal Circuit-focused Evaluation (\methodname) benchmark aims to extensively assess the professional capabilities of multimodal LLMs across all stages of the EDA circuit design workflow through closely-related QA samples.

\methodname~consists of 3614 manually curated test questions from diverse sources, covering general EDA knowledge and major circuit design stages for both digital and analog circuits. The questions specifically target different circuit-related abilities of multimodal LLMs.

Building on \methodname, we conduct extensive experiments to evaluate and horizontally compare the performance of existing LLMs in chip design, providing detailed analyses of their accuracies and errors.

\subsection{Data Collection and Curation}

\subsubsection{Data collection and organization}

To ensure the diversity and comprehensive coverage of \methodname~for fair comparisons, we collect and categorize the data according to the following three key aspects:

\textbf{Design stages.} Typical circuit design workflows are divided into front-end and back-end designs, based on predefined specifications. Our data collection spans all three stages of the design process.
\begin{itemize}
    \item \textit{Design specification.} To address the lack of systematic test questions related to design specifications, we gather 42 datasheets from online sources and generate high-quality questions focusing on specific parameters of the corresponding products with human experts review. The questions involve comprehension of specification documents, multi-step inference based on circuit design knowledge, and parameter-related computation.
    \item \textit{Front-end design.} We collect questions covering basic front-end circuit knowledge and code logic, netlist comprehension and computation, and circuit behavior analysis. Most questions are extracted from textbooks and online materials. Additionally, we manually craft 98 Verilog code snippets to test the models' RTL code comprehension capabilities.
    \item \textit{Back-end design.} We collect questions related to fundamental layout design knowledge and usages of layout design tools. The questions are extracted from textbooks and online materials.
    \item \textit{General knowledge.} 604 test questions related to general EDA knowledge (e.g.,~fundamentals of digital and analog circuits) are also included in the dataset.
\end{itemize}

\textbf{Circuit types.} We ensure a balanced data proportion between digital and analog circuit types. For \textit{specification} and \textit{back-end design}, we collect both digital and analog datasheets and layouts for test question extraction. For \textit{front-end design}, given that RTL codes are inherently digital, we specifically formulate 185 questions related to analog concepts (e.g,~voltage, current, and power calculations) to assess the models' analog computation capabilities and maintain data balance.

\textbf{Question types and modalities.} The questions in \methodname~can be categorized into single-answer and multi-answer choice questions, fill-in-the-blank questions, and open-ended questions with respect to question type, and can be categorized into text-only questions and multimodal questions with respect to data modality.

In addition, we paraphrase some of the question texts as basic question set augmentation. We also include explanations for each question-solution pair where applicable. This feature enables more objective and logical answer evaluation outlined in~\ref{sec:method_benchmark}.


\subsubsection{Data curation} To ensure the correctness and the quality of the collected questions, answers and explanations, our team, with specialized EDA engineers and Ph.D. students, manually review each question-solution pair in the dataset according to the following systematic set of rules:
\begin{itemize}
    \item \textbf{Relevance}: whether the question is strictly circuit-focused;
    \item \textbf{Alignment}: whether the question fits within the designated category;
    \item \textbf{Quality}: whether the question is of low quality;
    \item \textbf{Content correctness}: whether the solution and explanation (if applicable) are correct;
    \item \textbf{Attribute correctness}: whether the question type, circuit type, and difficulty level of the question are correct and objective;
    \item \textbf{Format}: whether the question adheres to the pre-defined model-friendly format.
\end{itemize}

We filter out irrelevant questions and low-quality questions that:
\begin{itemize}
    \item Are over-simple;
    \item Lack essential information (e.g.,~figures of document pages in \textit{design specification});
    \item Need to be answered in figure (e.g.,~layout sketching in \textit{back-end design});
    \item Are beyond the capacity of existing large vision-language models or cannot be fairly evaluated (e.g.,~those with extremely high complexity or subjectivity).
\end{itemize}
After filtering, we correct false information in the remaining questions to ensure accuracy, relevance and technical depth.

In addition, we utilize GPT-4o to assign a \textit{tested ability} label to each question, which reflects the specific ability that the question assesses in the models. The possible labels include:
\begin{itemize}
    \item \textbf{Knowledge-related}, indicating whether the models possess the necessary circuit-related knowledge stored in their memory.
    \item \textbf{Comprehension-related}, indicating whether the models can extract relevant information from the provided material in the question.
    \item \textbf{Reasoning-related}, indicating whether the models can perform logical inference based on the given context.
    \item \textbf{Computation-related}, indicating whether the models can apply correct formulas to derive correct numerical results.
\end{itemize}
In our benchmark preparation, all contextual outputs produced by LLMs (e.g. the label assignment process) are followed by manual verification and correction to ensure accuracy.

The detailed data composition is presented in Table~\ref{tab:data}. Note that a very small proportion of the data cannot be categorized as either digital or analog. In addition, we do not categorize simple multimodal questions into \textit{general knowledge}. Instead, we consider them as simple and straightforward cases within their respective circuit design stages (e.g.,~gate circuits as simple gate-level netlists in \textit{front-end design}, analog circuits as basic cases in analog \textit{front-end design}, etc). This results in no multimodal question of category \textit{general knowledge}.

\begin{table}[!t]
\renewcommand{\arraystretch}{1.1}
\caption{Data statistics of \methodname.}
\label{tab:data}
\begin{center}
\begin{tabular}{cc}
\toprule
Data compositions & Statistics \\
\midrule
Categories (stages) & 4 \\
Total & 3614 \\
General knowledge & 604 (16.7\%) \\
Specification & 490 (13.6\%) \\
Front-end & 1516 (41.9\%) \\
Back-end & 1004 (27.8\%) \\
\midrule
Total (digital : analog) & 50.0\% : 50.0\% \\
General knowledge (digital : analog) & 50.0\% : 50.0\% \\
Specification (digital : analog) & 52.8\% : 47.2\% \\
Front-end (digital : analog) & 46.0\% : 54.0\% \\
Back-end (digital : analog) & 54.9\% : 45.1\% \\
\midrule
Total (text-only : multimodal) & 58.4\% : 41.6\% \\
General knowledge (text-only : multimodal) & 100.0\% : 0.0\% \\
Specification (text-only : multimodal) & 50.0\% : 50.0\% \\
Front-end (text-only : multimodal) & 50.0\% : 50.0\% \\
Back-end (text-only : multimodal) & 50.0\% : 50.0\% \\
\midrule
Single-answer choice & 738 (20.4\%) \\
Multi-answer choice & 86 (2.4\%) \\
Fill-in-the-blank & 396 (11.0\%) \\
Open-ended & 2394 (66.2\%) \\
\midrule
Knowledge-related & 1446 (40.0\%) \\
Comprehension-related & 410 (11.3\%) \\
Reasoning-related & 832 (23.0\%) \\
Computation-related & 926 (25.6\%) \\
\midrule
With solution explanation & 2271 (62.8\%) \\
Easy : Medium : Hard & 15.2\% : 58.7\% : 26.1\% \\
\bottomrule
\end{tabular}
\end{center}
\end{table}

\begin{table*}[!t]
\caption{Statistical comparison between \methodname~and other circuit-focused benchmarks. SC, MC, B, O of ``question types'' refer to single-answer choice, multi-answer choice, fill-in-the-blank, and open-ended.}
\label{tab:benchmark_comparison}
\vspace{-5pt}
\begin{center}
\begin{tabular}{cccccc}
\toprule
Benchmarks & Size & Categories & Modalities & Question
types & Sources \\
\midrule
\textbf{\methodname} & 3614 & 4 & Text-only, text\&image & SC, MC, B, O & Textbook, Internet, Handcraft, Synthesis \\
EDA Corpus~\cite{wu2024eda} & 1533 & 2 & Text-only & O & Handcraft \\ 
ORD-QA~\cite{pu2024customized} & 90 & 4 & Text-only & O & Synthesis \\
ChatICD-Bench~\cite{xu2024chipexpert} & 622 & 7 & Text-only & O & Textbooks, Internet, Handcraft \\
\bottomrule
\end{tabular}
\end{center}
\vspace{-10pt}
\end{table*}

\subsection{Benchmark Construction}\label{sec:method_benchmark}

We select a broad series of text-only and multimodal LLMs, and test their performance on the curated dataset. We comprehensively assess their average accuracies across different circuit types, design stages, tested abilities, data modalities, question types, and difficulty levels.

The evaluation of each question is based on the similarity between the provided solution and the model output. Since they are both text-only, we employ four prevalent text-centered metrics to measure their similarity:
\begin{itemize}
    \item Bilingual evaluation understudy (BLEU) score, which estimates the ratio of the output phrases existing in the solution. In \methodname, BLEU score is used to evaluate whether the model produces misleading information or hallucination. We specifically adopt the 4-gram BLEU score in our settings.
    \item Recall-oriented understudy for gisting evaluation (ROUGE) score, which estimates the ratio of the solution phrases existing in the model output. In \methodname, ROUGE score is used to evaluate whether the model output contains all the critical information needed to answer the question. We utilize the average of the 1-gram, 2-gram and longest common subsequence ROUGE scores in our settings.
    \item Embedding cosine similarity, which estimates the semantic consistency between the two answers. In \methodname, embedding cosine similarity is used to evaluate whether the model output follows a reasoning process logically similar to the solution. We employ the \texttt{Text-Embedding-3-Large} model provided by OpenAI in our settings.
    \item GPT preference. In \methodname, we leverage GPT-4-turbo~\cite{achiam2023gpt}, proficient in text processing, for overall correctness rating from a well-trained expert's perspective.
\end{itemize}
We assign a weight of 2 for GPT preference and a weight of 1 for others in the \methodname~evaluation. Additionally, we require the tested models to provide an explanation alongside each answer to facilitate better correctness judgment. As validated in section~\ref{sec:exp_validation}, an integration of the metrics is qualified for answer similarity evaluation in circuit-specific scenarios, even though GPT may not always perform well in providing correct answers.

\subsection{Highlights of \methodname}

A statistical comparison between \methodname~and existing circuit-focused benchmarks is presented in Table~\ref{tab:benchmark_comparison}. We highlight the key advantage of \methodname~over existing circuit-focused benchmarks:
\begin{itemize}
    \item \textbf{Large data volume and broad data spectrum.} According to our knowledge, \methodname~is the first benchmark that encompasses different stages of typical circuit design workflows. It also covers diverse circuit types, question types, and data modalities.
    \item \textbf{Comprehensive evaluation.} Fine-grained data categorization and tested ability assignment enable \methodname's multi-dimensional model performance evaluation, and also allow for a more nuanced understanding of LLM capabilities.
    \item \textbf{High scalability.} Usages of GPT with manual data curation achieves high efficiency and scalability of data collection without compromising data quality. Furthermore, GPT-generated questions exhibit minimal overlap with the training corpora of existing foundation models, ensuring fair horizontal comparisons.
\end{itemize}

\section{Experiments}\label{sec:exp}

\subsection{Baseline Models}\label{sec:exp_baseline}

We evaluate various model families that are widely-applied in the field of multimodal QA. For many model families with multiple variants and parameter scales, we select at least two variants from each to test their circuit-focused upscaling capabilities. These models can be categorized based on their image processing techniques:
\begin{itemize}
    \item Text-only, which lack the ability to process visual data. For these models, we leverage a BLIP~\cite{li2022blip} captioning model to generate textual descriptions as substitutes for visual information.
    \item Image-to-string, which process visual information through hard-coded image-to-string conversions.
    \item Image encoding, which incorporate embedded visual encoders to directly process images.
\end{itemize}
The list of models evaluated is presented in Table~\ref{tab:results}. According to our knowledge, ChipExpert~\cite{xu2024chipexpert} is currently the only publicly available circuit-focused LLM in this domain.

\subsection{Correctness Validation of the Proposed Evaluator}\label{sec:exp_validation}

We first conduct experiments to validate the effectiveness of our proposed comprehensive answer evaluation metric stated in Section~\ref{sec:method_benchmark}. We randomly select 100 questions from each design stage across multiple models and obtain multiple groups of answers. Our team then manually check whether the overall scores calculated with the proposed metric are accurate enough to reflect their correctness. Results show that the testers are generally positive about the quantitative results, indicating the effectiveness of the proposed evaluator.

\subsection{Evaluation Results and Discussions}\label{sec:exp_results}

\begin{table*}[!t]
\footnotesize
\renewcommand{\arraystretch}{1.19}
\caption{Comprehensive evaluation results of the selected text-only and multimodal LLMs. Among the model categories, the highest correctness is \textbf{bold}, and the second-highest is \underline{underlined}. The higher value ($\uparrow$) indicates better performance.}
\label{tab:results}
\begin{center}
\begin{tabular}{cc|cccc|cccc|cc}
\toprule
Models & Overall & G & S & F & B & K & Cph & R & Cpt & T & M \\
\midrule
\multicolumn{12}{c}{\textbf{MLLMs (Image encoding)}} \\
\midrule
QWen-VL-Chat~\cite{bai2023qwen} & 32.2 & 41.6 & 41.0 & 32.7 & 21.4 & 37.3 & 46.4 & 25.6 & 16.9 & 34.7 & 28.6 \\
InternLM-XComposer-VL-7B~\cite{zhang2023internlm} & 19.7 & 41.2 & 11.9 & 22.9 & 5.8 & 32.5 & 17.0 & 15.2 & 12.5 & 25.5 & 11.6 \\
InternVL2-8B~\cite{chen2024internvl} & 38.8 & \underline{48.3} & 34.9 & 45.5 & 24.8 & 51.8 & 47.1 & 39.5 & 29.0 & 50.3 & 22.5 \\
InternVL2-40B~\cite{chen2024internvl} & 41.7 & \underline{48.3} & 35.4 & 51.8 & \underline{25.7} & \underline{54.4} & 51.8 & \underline{45.7} & \underline{30.3} & \underline{54.3} & 24.2 \\
InstructBLIP-Flan-T5-XL~\cite{dai2023instructblip} & 10.8 & 11.9 & 18.9 & 8.6 & 9.6 & 8.2 & 16.5 & 6.8 & 7.1 & 8.1 & 14.6 \\
InstructBLIP-Flan-T5-XXL~\cite{dai2023instructblip} & 10.7 & 17.7 & 11.1 & 6.7 & 12.2 & 11.6 & 13.2 & 9.5 & 3.3 & 9.8 & 11.9 \\
BLIP2-Flan-T5-XL~\cite{li2023blip} & 14.3 & 14.4 & 20.3 & 13.0 & 13.2 & 12.4 & 24.2 & 12.4 & 7.7 & 13.8 & 15.0 \\
BLIP2-Flan-T5-XXL~\cite{li2023blip} & 14.0 & 14.1 & 17.3 & 12.9 & 13.9 & 14.8 & 18.8 & 15.0 & 9.8 & 15.6 & 11.7 \\
LlaMa3.2-Vision-Instruct-11B~\cite{dubey2024llama} & \underline{43.2} & 42.1 & \underline{58.0} & \underline{53.5} & 21.1 & 38.9 & \underline{54.5} & 42.7 & 29.9 & 39.7 & \underline{48.1} \\
LlaMa3.2-Vision-Instruct-90B~\cite{dubey2024llama} & \textbf{58.5} & \textbf{64.2} & \textbf{63.6} & \textbf{69.1} & \textbf{36.6} & \textbf{62.2} & \textbf{71.4} & \textbf{57.0} & \textbf{42.7} & \textbf{61.2} & \textbf{54.7} \\
MiniCPM-V~\cite{yao2024minicpm} & 21.1 & 23.0 & 19.2 & 25.9 & 13.6 & 24.5 & 28.7 & 22.5 & 11.3 & 24.5 & 16.3 \\
MiniCPM-V2~\cite{yao2024minicpm} & 11.5 & 17.7 & 13.3 & 11.8 & 6.6 & 14.2 & 15.5 & 7.8 & 0.7 & 10.5 & 13.1 \\
MiniCPM-LlaMa3-V2.5~\cite{yao2024minicpm} & 43.0 & 46.7 & 47.4 & 52.5 & 24.3 & 44.5 & 53.5 & 44.5 & 25.6 & 43.7 & 41.9 \\
Yi-VL-6B~\cite{ai2024yi} & 17.3 & 30.0 & 12.9 & 20.2 & 7.4 & 23.8 & 19.3 & 15.4 & 6.4 & 19.5 & 14.1 \\
Yi-VL-34B~\cite{ai2024yi} & 32.5 & 34.4 & 40.6 & 38.9 & 17.6 & 34.1 & 45.8 & 28.9 & 16.8 & 32.9 & 31.8 \\
Kosmos2~\cite{peng2023kosmos} & 13.6 & 10.9 & 15.3 & 13.7 & 14.1 & 13.7 & 15.1 & 13.3 & 9.4 & 12.9 & 14.5 \\
\midrule
\multicolumn{12}{c}{\textbf{MLLMs (Image-to-string)}} \\
\midrule
GPT-4~\cite{achiam2023gpt} & 63.7 & 64.7 & 63.8 & 74.0 & 47.6 & 66.7 & 69.5 & 61.3 & 45.2 & 62.0 & 66.1 \\
GPT-4-Turbo~\cite{achiam2023gpt} & 67.4 & 67.9 & 68.9 & \underline{77.9} & \textbf{50.6} & \textbf{69.4} & 74.0 & \underline{66.7} & 54.0 & \textbf{67.2} & 67.8 \\
GPT-4v~\cite{achiam2023gpt} & \textbf{69.4} & \textbf{69.9} & \underline{80.3} & \textbf{79.8} & 48.2 & \underline{67.9} & \underline{75.5} & \textbf{67.3} & \textbf{59.2} & \underline{66.5} & \textbf{73.6} \\
GPT-4o~\cite{hurst2024gpt} & \underline{68.0} & \underline{69.4} & \textbf{80.7} & 76.2 & \underline{48.6} & 66.1 & \textbf{75.9} & 66.3 & \underline{56.9} & 65.5 & \underline{71.4} \\
Reka-Flash~\cite{team2024reka} & 55.7 & 61.8 & 54.7 & 68.3 & 33.5 & 63.0 & 66.5 & 61.3 & 39.3 & 63.4 & 44.8 \\
Reka-Edge~\cite{team2024reka} & 36.7 & 43.4 & 35.4 & 45.8 & 19.5 & 42.5 & 46.2 & 41.9 & 19.4 & 42.0 & 29.3 \\
\midrule
\multicolumn{12}{c}{\textbf{Text-only LLMs}} \\
\midrule
ChipExpert*~\cite{xu2024chipexpert} & \textbf{67.1} & \textbf{75.5} & \textbf{61.9} & 69.5 & \textbf{61.1} & \textbf{77.4} & \textbf{77.0} & \textbf{69.1} & \textbf{63.7} & \textbf{77.7} & \textbf{52.4} \\
GPT-3.5-Turbo~\cite{ouyang2022training} & 54.9 & 51.7 & 58.6 & 66.7 & 37.2 & 59.6 & 66.4 & 58.4 & 42.1 & 60.8 & 46.7 \\
DeepSeek-MoE-16B-Chat~\cite{dai2024deepseekmoe} & 34.0 & 28.0 & 34.6 & 36.7 & 33.3 & 33.2 & 38.8 & 34.8 & 34.5 & 35.6 & 31.8 \\
DeepSeek-LLM-7B-Chat~\cite{bi2024deepseek} & 35.1 & 27.4 & 39.4 & 37.1 & 34.7 & 33.5 & 42.8 & 36.4 & 34.3 & 36.2 & 33.5 \\
DeepSeek-Math-7B-RL~\cite{shao2024deepseekmath} & 39.9 & 42.3 & 42.8 & 41.6 & 34.3 & 42.6 & 44.9 & 39.8 & 42.7 & 44.1 & 33.9 \\
DeepSeek-Math-7B-Instruct~\cite{shao2024deepseekmath} & 40.5 & 53.3 & 40.1 & 41.0 & 32.3 & 46.2 & 42.8 & 38.3 & 46.0 & 47.4 & 30.8 \\
DeepSeek-V2-Lite-Chat~\cite{liu2024deepseek} & 39.7 & 35.0 & 40.7 & 43.1 & 36.8 & 40.4 & 44.3 & 41.2 & 40.1 & 42.6 & 35.5 \\
QWen-2-Instruct-0.5B~\cite{yang2024qwen2} & 13.8 & 13.6 & 23.1 & 14.4 & 8.3 & 13.9 & 25.8 & 11.1 & 6.9 & 15.3 & 11.6 \\
QWen-2-Instruct-7B~\cite{yang2024qwen2} & 48.2 & 54.3 & 51.2 & 59.5 & 26.0 & 55.2 & 58.2 & 51.3 & 30.4 & 53.8 & 40.3 \\
QWen-2-Instruct-72B~\cite{yang2024qwen2} & 50.2 & 66.4 & 37.6 & 59.8 & 32.3 & 64.0 & 45.2 & 53.1 & 40.7 & 58.3 & 39.0 \\
QWen-2.5-Instruct-7B~\cite{yang2024qwen2} & 53.0 & 57.0 & 57.4 & 63.3 & 32.9 & 58.9 & 65.9 & 54.0 & 38.9 & 59.1 & 44.4 \\
QWen-2.5-Instruct-72B~\cite{yang2024qwen2} & 60.9 & 64.2 & \textbf{61.9} & \underline{73.3} & 39.7 & 66.0 & \underline{72.3} & 64.3 & \underline{49.7} & 67.5 & \underline{51.6} \\
InternLM-Chat-20B~\cite{team2023internlm} & 33.3 & 35.9 & 38.9 & 39.0 & 20.5 & 37.7 & 42.6 & 34.6 & 20.2 & 37.2 & 27.8 \\
InternLM2-Chat-7B~\cite{team2023internlm} & 45.0 & 53.0 & 49.7 & 52.7 & 26.3 & 50.3 & 55.5 & 48.5 & 32.3 & 51.1 & 36.4 \\
InternLM2.5-Chat-7B~\cite{team2023internlm} & 46.7 & 57.5 & 51.0 & 55.3 & 25.0 & 54.0 & 60.5 & 45.1 & 33.6 & 53.4 & 37.3 \\
LlaMa2-Chat-HF-7B~\cite{touvron2023llama} & 29.7 & 37.7 & 31.3 & 35.8 & 14.8 & 35.5 & 40.3 & 29.9 & 12.5 & 33.6 & 24.2 \\
LlaMa2-Chat-HF-13B~\cite{touvron2023llama} & 33.6 & 42.9 & 38.8 & 38.0 & 18.9 & 39.5 & 48.1 & 33.4 & 11.2 & 37.3 & 28.5 \\
LlaMa3-Instruct-8B~\cite{dubey2024llama} & 46.1 & 51.7 & 47.2 & 55.6 & 27.8 & 53.1 & 59.2 & 47.5 & 31.1 & 53.0 & 36.3 \\
LlaMa3.1-Instruct-8B~\cite{dubey2024llama} & 47.6 & 53.0 & 50.6 & 57.8 & 27.5 & 54.8 & 60.7 & 50.3 & 35.9 & 56.2 & 35.6 \\
MiniCPM-SFT-1B~\cite{hu2024minicpm} & 20.5 & 31.0 & 22.5 & 20.7 & 12.9 & 26.6 & 27.5 & 15.1 & 6.1 & 21.9 & 18.6 \\
MiniCPM-SFT-2B~\cite{hu2024minicpm} & 24.8 & 29.1 & 23.6 & 26.9 & 19.6 & 29.2 & 31.2 & 22.8 & 10.9 & 26.1 & 23.0 \\
MiniCPM3-4B~\cite{hu2024minicpm} & 48.6 & 55.1 & 49.3 & 58.0 & 30.3 & 53.7 & 61.6 & 50.9 & 31.6 & 53.7 & 41.4 \\
Yi-Chat-6B~\cite{ai2024yi} & 31.7 & 41.8 & 35.9 & 35.6 & 17.6 & 37.4 & 45.7 & 31.8 & 11.4 & 35.7 & 26.1 \\
ChatGLM3-6B~\cite{glm2024chatglm} & 32.2 & 39.6 & 37.4 & 37.8 & 16.7 & 36.9 & 46.6 & 32.7 & 12.3 & 35.9 & 27.0 \\
Gemini1.0-Pro$^+$~\cite{team2023gemini} & 18.8 & 41.1 & 1.9 & 9.0 & 28.6 & 34.7 & 8.5 & 15.4 & 16.4 & 28.6 & 5.1 \\
Gemini1.5-Pro$^+$~\cite{team2024gemini} & \underline{62.2} & \underline{72.2} & 50.2 & \textbf{75.2} & \underline{42.6} & \underline{75.1} & 66.0 & \underline{65.3} & 45.2 & 70.6 & 50.4 \\
Claude3.5-Sonnet$^+$~\cite{claude2024claude} & 53.5 & 60.6 & 35.9 & 65.4 & 39.9 & 74.4 & 62.5 & 54.4 & 40.0 & \underline{70.8} & 29.3 \\
\bottomrule
\end{tabular}
\begin{tablenotes}
\centering
\footnotesize
\item[1] {\bf *} A circuit-focused LLM based on the LlaMa3-8B architecture, trained with circuit-related curated data.
\item[2] {\bf +} Multimodal models that only support text-only massive testing.
\end{tablenotes}
\end{center}
\vspace{-10pt}
\end{table*}

\begin{table*}[!t]
\caption{Average evaluation results of the selected text-only and multimodal LLMs of each model category.}
\label{tab:avg_results}
\vspace{-5pt}
\begin{center}
\begin{tabular}{cc|cccc|cccc|cc}
\toprule
Model Categories & Overall & G & S & F & B & K & Cph & R & Cpt & T & M \\
\midrule
MLLMs (Image encoding) & 26.4 & 31.7 & 28.9 & 30.0 & 16.6 & 29.9 & 33.7 & 25.1 & 16.2 & 28.6 & 23.4\\
MLLMs (Image-to-string) & \textbf{60.0} & \textbf{62.9} & \textbf{65.2} & \textbf{69.7} & \textbf{41.1} & \textbf{62.4} & \textbf{67.8} & \textbf{60.6} & \textbf{45.5} & \textbf{60.9} & \textbf{58.7} \\
Text-only LLMs & 41.0 & 47.1 & 40.7 & 46.8 & 28.7 & 47.3 & 49.5 & 41.6 & 30.2 & 46.6 & 33.1 \\
\bottomrule
\end{tabular}
\end{center}
\vspace{-10pt}
\end{table*}

The overall evaluation results of the models are detailed in Table~\ref{tab:results}, with the average results of each model category shown in Table~\ref{tab:avg_results}. The baseline models are tested and horizontally compared across the following aspects:
\begin{itemize}
    \item \textbf{Global baseline performance (Overall)}, comprehensively reflecting the models' performance on circuit-related QA;
    \item \textbf{Circuit design stages (G / S / F / B)}, targeting the models' knowledge base and logic capabilities in specific circuit design scenarios. In the table, G, S, F, and B respectively refer to general knowledge, design specification, front-end design, and back-end design stages;
    \item \textbf{Tested abilities (K / Cph / R / Cpt)}, indicating whether the models' original capabilities on general-purpose benchmarks maintain effectiveness in the EDA field. In the table, K, Cph, R, and Cpt respectively represent that the questions test the abilities of knowledge, comprehension, reasoning, and computation;
    \item \textbf{Data modalities (T / M)}, indicating whether the models can effectively read and process circuit-related visual materials. In the table, T and M respectively refer to text-only and multimodal questions.
\end{itemize}

\textbf{Global baseline performance (Overall).} From this table, we observe that most of LLMs struggle to reach satisfactory levels of accuracy on the overall problem set. For instance, several LLMs, such as InstructBLIP~\cite{dai2023instructblip} and BLIP2~\cite{li2023blip} only perform less than 20\% accuracy in our~\methodname. The primary reason for this shortfall is the lack of sufficient circuit-related training materials in existing vision-language corpora. In addition, due to the scarcity of circuit-specific data in LLM training, the circuit-focused horizontal performance between them may not align well with general-purpose benchmarks. Nonetheless, most models exhibit certain scalability as previously validated by other benchmarks.

Among the evaluated models, GPT-4v~\cite{achiam2023gpt} achieves the highest overall performance with 69.4\% problem solved out, while ChipExpert~\cite{xu2024chipexpert} ranks highest among open-source models by answering 67.1\% questions correctly. Models such as LlaMa3.2-Vision-Instruct-90B~\cite{dubey2024llama} (58.5\%) and Gemini1.5-Pro~\cite{team2024gemini} (62.2\%) also show relatively better performance than most other LLMs. Notably, these models all have significantly large parameter scales except ChipExpert, indicating their high scalability, and that we can improve their performance in circuit design simply by upscaling. As a circuit-focused model, the text-only ChipExpert, based on the LlaMa3-8B~\cite{dubey2024llama} architecture, outperforms most other models by a large margin, particularly in knowledge mastery and back-end processing, further highlighting the significance of high-quality training data in circuit-related LLM applications.

\textbf{Performance across different circuit design stages (G / S / F / B).} We further investigate the results according to the different design stages. The results are evident that back-end design generally exhibits the lowest accuracy across models. For example, GPT-4v~\cite{achiam2023gpt} achieves a relatively high accuracy of 69.4\% on overall questions, but its performance on back-end design questions is considerably lower, with only 48.2\% accuracy. This trend is mostly consistent across other models, with ChipExpert~\cite{xu2024chipexpert} answering 61.1\% of the back-end design questions correctly, while models like LlaMa3.2-Vision-Instruct-90B~\cite{dubey2024llama} and Gemini1.5-Pro~\cite{team2024gemini} only manage 36.6\% and 42.6\%, respectively. On average, different categories of LLMs have a 12.0\% to 21.8\% performance decline on back-end problems compared with those of other circuit design stages.

The disparity in performance can be attributed to several key factors. First, questions related to general knowledge (e.g.,~basic digital and analog circuits) and front-end design (e.g., comprehension of code snippets and simple circuit diagrams) are more prevalent in the existing training corpora, making it easier for LLMs to handle these tasks. In contrast, back-end design questions often involve highly specific layouts and details that require more specialized data, which is currently scarce in most circuit-related corpora used to train LLMs. Second, back-end design is more context-dependent and multimodal, with many questions requiring a deep understanding of placement, routing, and other intricate layout-specific challenges. These complex visual and spatial relationships pose a significant difficulty for LLMs that have been trained primarily on textual data with limited circuit-specific examples. 


\textbf{Performance across different tested abilities (K / Cph / R / Cpt).} The evaluation results highlight distinct performance patterns of LLMs across different abilities. Specifically, most models perform strongly on general knowledge retrieval (K) and basic comprehension (Cph) tasks. For instance, GPT-4o~\cite{hurst2024gpt} achieves an impressive accuracy of 75.9\% on comprehension (Cph) tasks, while GPT-4-Turbo~\cite{achiam2023gpt} successfully answers 69.4\% of knowledge retrieval (K) questions. However, there is a noticeable decline in performance when models are tasked with more complex reasoning and computation problems. For example, GPT-4o~\cite{hurst2024gpt} solves only 56.9\% of computation-related (Cpt) questions, and several other LLMs, including InstructBLIP~\cite{dai2023instructblip}, MiniCPM~\cite{hu2024minicpm}, and InternLM~\cite{team2023internlm}, answer less than 50\% of reasoning and computation questions correctly. On average, different categories of LLMs have a 1.8\% to 5.7\% performance decline on reasoning compared with knowledge retrieval and comprehension, and an 8.9\% to 15.1\% further performance decline on computation.

The underlying reasons for this performance disparity are twofold. First, the LLMs are primarily trained on large vision-language corpora, which are designed to enhance general comprehension and knowledge retrieval. Second, while LLMs excel in general reasoning tasks and numerical computations~\cite{yue2024mmmu}, reasoning and computing circuit-based problems involve specific electronic rules and design methodology. This domain-specific knowledge gap leads to lower accuracy in reasoning and computation tasks within the circuit design context. 


\textbf{Performance across different data modalities (T / M).} Evaluation results also highlight notable differences in performance when LLMs are tasked with processing circuit-related images. Although multimodal LLMs incorporate image encoders to encode or summarize non-textual information, most models experience performance degradation on multimodal problems (M) compared to purely text-only problems (T). For example, LlaMa3.2-Vision-Instruct-90B~\cite{dubey2024llama} answers 61.2\% text-only questions but answers 54.7\% with multimodal statements. On average, different categories of LLMs have a 2.2\% to 13.5\% performance decline on multimodal problems compared with text-only ones. However, the GPT model family achieve better accuracy on multimodal questions than text-only ones. The image-to-string conversion does not result in severe information loss, which can better cooperate with the well-trained GPT backbones. More importantly, images in string format can be regarded as text and processed in a similar way to the question texts, thus avoiding the accumulation of errors by the visual encoders.

Interestingly, we also find that most models with visual encoders generally perform even worse than text-only models in multimodal QA, given that well-trained visual encoders are supposed to extract more accurate and comprehensive information than short image captions. For instance, apart from the LlaMa3-backboned models (i.e., LlaMa3.2-Vision-Instruct-11B~\cite{dubey2024llama}, LlaMa3.2-Vision-Instruct-90B~\cite{dubey2024llama}, and MiniCPM-LlaMa3-V2.5~\cite{yao2024minicpm}), the models integrated with visual encoders have the highest multimodal QA accuracy 31.8\% (achieved by Yi-VL-34B~\cite{ai2024yi}), which is lower than 55.6\% of the text-only models tested.
One possible major reason is that the embedded image encoders are not specifically trained with circuit-related data, and may produce false visual embeddings that mislead the backbone LLMs and negatively impact their final outputs.

\subsection{Possible Model Improvements for Circuit-Related Tasks}

\textbf{How should we process circuit-related images?} As stated above, visual encoders that are not specifically trained with circuit-related data may lead to even worse performance than leveraging image captions. Meanwhile, there is currently no circuit-focused visual encoder that matches the performance of general-purpose ones while adequately covering all circuit design stages. However, embeddings generated by well-trained visual encoders have much more information than short captions. Therefore, with sufficient time and computation resources, visual encoders are apparently better than captioning. On the other hand, encoding images into strings is also a satisfactory approach, as it does not require extra visual processors and does not suffer from severe information loss. However, there is a high proportion of redundant and irrelevant character tokens in the encoded strings, which may severely increase the burden on the subsequent LLM backbone. However, this problem can also be solved by a powerful LLM backbone, which is proved by the high performance of the GPT-4 model family.

\textbf{What are the effectiveness and costs of model re-training?} Most tested models are capable of processing general questions with high accuracies, as validated by other general-purpose benchmarks. With the same parameter scales and the problems restricted within the circuit-related field, it is evident that the models have the potential capabilities to reach equally high or even better performance. This is empirically validated by the performance of ChipExpert~\cite{xu2024chipexpert}, which demonstrates that even a small-sized LLM can achieve relatively high circuit-related performance with the help of targeted training.

However, in the circuit-related field, the major challenge of large model training is not the choice of backbone models, but the collection of relevant high-quality data. Most large-scale general-purpose training corpora~\cite{yang2018hotpotqa,rajpurkar2016squad} are established based on open-sourced online materials with clear in-context answers. However, high-quality, open-sourced, circuit-related materials are extremely scarce, reflected by the fact that the scales of existing circuit-related benchmarks are generally only 10\% of the general-purpose ones. This poses a great challenge for LLM training for related downstream circuit-related tasks. However, with sufficient high-quality data, it is highly possible for existing LLMs to achieve much higher performance than that reported in the paper.

The time and computation resources required for model training and fine-tuning are the same as training on general-purpose data. From the multimodal perspective, image-to-string models require re-training or fine-tuning the whole backbone, while models with visual encoders can adopt specific techniques to achieve fine-tuning only on the visual encoders. Ultimately, the choice of image processing strategies determines the required resources and difficulty in model training and fine-tuning.

\textbf{How to improve the models' test-time performance?} There are multiple test-time techniques to improve the performance of LLMs under limited resources. In this paper, we adopt the Chain-of-Thought (CoT) reasoning, a widely-adopted test-time inference technique~\cite{brown2020language} as the core improvement approach. CoT encourages the model to break down complex reasoning tasks into smaller, more manageable steps, effectively improving the model's ability to handle intricate problems. By generating intermediate reasoning steps, CoT improves the model's decision-making process and overall accuracy.

We select three representative LLMs, BLIP2-Flan-T5-XL~\cite{li2023blip} (image encoding), GPT-4o~\cite{hurst2024gpt} (image-to-string), and DeepSeek-LLM-7B-Chat~\cite{bi2024deepseek} (text-only) to comprehensively evaluate the impact of CoT. We randomly select 100 questions from each circuit design stage, and add detailed CoT instructions to the query prompts. The instructions include the following sequential steps:
\begin{enumerate}
    \item Understand and clarify the core of the question;
    \item Locate relevant information presented in the material;
    \item Identify and extract key data crucial to correct solutions;
    \item Apply relevant knowledge and logic chain to solve the question;
    \item Self-check the correctness and consistency of the generated answer;
    \item Summarize the generated content and output the final answer with explanations.
\end{enumerate}

\begin{table}[!t]
\caption{Performance comparison of the selected models before and after CoT integration. Among the selected models, the higher correctness is \textbf{bold}.}
\label{tab:cot_results}
\vspace{-5pt}
\begin{center}
\resizebox{\linewidth}{85pt}{
\begin{tabular}{lcc|cccc}
\toprule
Models & & Overall & G & S & F & B \\
\midrule
BLIP2~\cite{li2023blip} w/ CoT & & \textbf{17.3} & \textbf{14.9} & 20.2 & \textbf{22.2} & 12.0 \\
BLIP2~\cite{li2023blip} w/o CoT& & 14.3 & 14.4 & \textbf{20.3} & 13.0 & \textbf{13.2} \\
\midrule
GPT-4o~\cite{hurst2024gpt} w/ CoT & & \textbf{68.6} & 67.7 & 80.2 & \textbf{79.9} & \textbf{49.5} \\
GPT-4o~\cite{hurst2024gpt} w/o CoT & & 68.0 & \textbf{69.4} & \textbf{80.7} & 76.2 & 48.6 \\
\midrule
DeepSeek~\cite{bi2024deepseek} w/ CoT & & \textbf{36.4} & \textbf{29.5} & 38.9 & \textbf{40.3} & \textbf{37.0} \\
DeepSeek~\cite{bi2024deepseek} w/o CoT & & 35.1 & 27.4 & \textbf{39.4} & 37.1 & 34.7 \\
\midrule
Models & T & M & K & Cph & R & Cpt \\
\midrule
BLIP2~\cite{li2023blip} w/ CoT & \textbf{20.9} & \textbf{15.5} & \textbf{15.7} & \textbf{25.7} & \textbf{16.9} & \textbf{9.1} \\
BLIP2~\cite{li2023blip} w/o CoT & 13.8 & 15.0 & 12.4 & 24.2 & 12.4 & 7.7 \\
\midrule
GPT-4o~\cite{hurst2024gpt} w/ CoT & 65.0 & \textbf{73.4} & 64.4 & 75.9 & \textbf{71.1} & \textbf{59.8} \\
GPT-4o~\cite{hurst2024gpt} w/o CoT & \textbf{65.5} & 71.4 & \textbf{66.1} & 75.9 & 66.3 & 56.9 \\
\midrule
DeepSeek~\cite{bi2024deepseek} w/ CoT & \textbf{37.3} & \textbf{35.6} & 32.9 & 42.4 & \textbf{37.1} & \textbf{34.8} \\
DeepSeek~\cite{bi2024deepseek} w/o CoT & 36.2 & 33.5 & \textbf{33.5} & \textbf{42.8} & 36.4 & 34.3 \\
\bottomrule
\end{tabular}
}
\end{center}
\vspace{-10pt}
\end{table}

With the CoT technique, the three models have achieved different degrees of performance improvements, with detailed numerical results presented in Table~\ref{tab:cot_results}. BLIP2-Flan-T5-XL~\cite{li2023blip}, GPT-4o~\cite{hurst2024gpt}, and DeepSeek-LLM-7B-Chat~\cite{bi2024deepseek} have 3.0\%, 0.6\%, and 1.3\% overall performance increases, respectively. In terms of circuit design stages, the models are most significantly improved on front-end QA, with BLIP2-Flan-T5-XL~\cite{li2023blip} achieving relatively the highest increase in correctness by 9.2\%. In terms of tested abilities, the models' reasoning abilities are most significantly improved. GPT-4o~\cite{hurst2024gpt} achieves relatively the highest increase in reasoning correctness by 4.8\%.

Questions of different circuit design stages and tested abilities result in different impacts of CoT on the tested models. Questions related to general knowledge retrieval require sufficient and high-quality knowledge bases recorded in the models' parameters. Therefore, CoT instructions on inference may not be remarkably effective. Questions of specifications and back-end design suffer from the same issue, resulted by the scarce relevant data in LLM training corpora. In contrast, front-end design materials are relatively the easiest to obtain from open sources, while many of the corresponding questions require certain inference abilities. Therefore, CoT has relatively the highest impact on front-end QA. In terms of tested abilities, CoT also has similar impact patterns. The requirements on logic inference abilities grow higher from knowledge retrieval to numerical computation. However, without knowledge of the correct formulas and algorithms to apply, it is still a great challenge to improve the models' computation abilities, even with the help of CoT. Therefore, CoT is more suitable for solving reasoning-related questions. On the other hand, low-performance models may not benefit much from CoT, as their accuracy is bottlenecked by the insufficient data and training processes.

In summary, from the conducted experiments, CoT is an effective approach of test-time model performance improvement on circuit-related QA, and is specifically effective for front-end questions with relatively high occurrence in the training data, but requiring high logic inference capabilities.
\section{Limitations}\label{Sec:Limitation}

\textbf{Usage of MLLMs for answer correctness evaluation.} \methodname~leverages GPT-4o~\cite{hurst2024gpt} for answer correctness evaluation. This may result in two issues: the evaluation may not be 100\% accurate; it may favor models in the same family (e.g., models in the GPT series in this paper). However, the large question corpus of \methodname~makes it practically impossible to conduct manual answer assessment. In addition, as indicated in Section~\ref{sec:exp_validation}, powerful MLLMs such as GPT-4o are generally qualified for this task.

\textbf{Data quality.} Sources of \methodname's questions include textbook, online corpora, handcraft, and basic augmentation-based synthesis from existing questions. Since errors may occur in question extraction, crafting, and augmentation, quality of the questions and correctness of the truth answers cannot be 100\% guaranteed. We conduct multiple rounds of manual checking to solve this issue to the most extent. The data quality of \methodname~can also be iteratively refined.

\section{Conclusion}\label{Sec:Conclusion}


In this paper, we present \methodname, a comprehensive circuit-focused benchmark for evaluating multimodal LLMs. With a wide spectrum of test questions covering scenarios such as general circuit knowledge, design specification, front-end design, and back-end design, \methodname~is capable of extensively evaluating general MLLMs on their circuit-related capabilities. In addition to horizontal comparisons, \methodname~highlights that existing efforts on large foundation models are generally underdeveloped in the circuit and EDA fields. Finally, we explore potential solutions and propose directions for future advancements. We believe \methodname~could foster collaboration between the AI and hardware communities and stimulate progress at the intersection of LLMs and circuit design.

\balance


\bibliographystyle{IEEEtran}
\bibliography{reference}

\end{document}